\documentclass[runningheads]{llncs}
\usepackage{graphicx}
\usepackage{amsmath,amssymb} 
\usepackage{algorithm}
\usepackage[noend]{algpseudocode}
\usepackage{color}
\usepackage[width=122mm,left=12mm,paperwidth=146mm,height=193mm,top=12mm,paperheight=217mm]{geometry}
\begin{document}
\pagestyle{headings}
\mainmatter

\title{Outline Colorization through Tandem Adversarial Networks.} 

\author{Kevin Frans}


\institute{Henry M. Gunn High School, Palo Alto, CA\\
	\email{ kevinfrans2@gmail.com}
}

\maketitle

\begin{abstract}
When creating digital art, coloring and shading are often time consuming tasks that follow the same general patterns. A solution to automatically colorize raw line art would have many practical applications. We propose a setup utilizing two networks in tandem: a color prediction network based only on outlines, and a shading network conditioned on both outlines and a color scheme. We present processing methods to limit information passed in the color scheme, improving generalization. Finally, we demonstrate natural-looking results when colorizing outlines from scratch, as well as from a messy, user-defined color scheme.
\end{abstract}

\section{Introduction}

In digital art, the creation process often starts with a black and white outline, which is then colored and shaded in, resulting in a final product. While drawing the initial lines requires an artist's creativity and imagination, shading (and to an extent, colorization) often follows the same patterns. Shadows, highlights, and gradients can be inferred from the structure of the line art. Automating this process has a multitude of practical applications: the production time of asset-heavy creations such as computer games can be cut drastically. Furthermore, the vast amounts of comics and manga available only in black and white can be converted into colorized versions.

Generative adversarial networks\cite{gan} has been shown to produce high quality images, even when there are multiple correct outputs for a single input, such as when coloring an arbitrary piece of clothing. However, training directly from outlines to colored images leads to misshapen results, often with similar colors placed in unrelated locations.

In this paper, we present an approach to generating colored and shaded artwork from outlines, using two distinct convolutional networks in tandem. We break the problem down into two easier ones, solved separately. First, we infer a general color scheme from an outline, in the color prediction network. Next, the shading network takes this color scheme, along with an outline, and produces a final image.

By structuring our networks in a tandem setup, the shading network's job is simplified to inserting already-known colors in the correct places. The color prediction network handles any ambiguities in assigning colors. To account for potential error in color prediction, we randomly remove information in the color scheme during training, forcing the shading network to generalize when given imperfect inputs.

\section{Related Work}

There have been many advances in the field of image to image transformation. Mirza et. al\cite{cgan} introduced a conditional variety of the generative adversarial network, where the data conditioned on is given to both the generator and the discriminator. The method achieved solid results on the tasks of generating MNIST digits, and tagging photos from the MIR Flickr 25,000 dataset. Isola et. al\cite{image2image} present an outline for mapping images to one another, by optimizing pixel loss as well as an adversarial loss function. They treat the discriminator similarly to a convolutional layer, allowing it to see patches as it is slid across the generated image.

More specifically, the problem of synthesizing realistic images from sketches has been the subject of past studies. Convolutional Sketch Inversion\cite{sketch} attempts to recreate photorealistic faces from sketches. They made use of a deep network of convolutional layers, trained directly using both pixelwise and feature loss. Both computer-generated datasets and hand-drawn sketches were utilized. PaintsChainer\cite{paintschainer} uses a more recent approach, training a conditional generative adversarial network to directly predict colors from line art, and utilizing color hints. Levin et. al\cite{coloroptim}, Iizuka et. al\cite{lettherebe}, Zhang et. al\cite{colorful}, and Larsson et.al\cite{learnrep} also address the task of image colorization.

Finally, there has been research into utilizing multiple networks in conjunction. StackGAN\cite{stackgan} proposes a setup involving two networks working in coordination, for the task of text-to-image synthesis. One network generates low-resolution images from captions, while another increases resolution and adds fine details.

\section{Network Structure}

\subsection{Fully Convolutional Network}

As our problem is essentially an image mapping task, our network will consist of many convolutional layers, followed by rectified linear units. We set the convolutional filter size to 5x5, with a stride of 2, so for each layer the height and width of our feature matrix decreases by a factor of two. In addition, we double the number of features every layer, starting from 64.

Once a dense feature matrix is reached, the network switches to transpose convolution layers, allowing it to increase its height and width, while condensing features into what will eventually be three color channels. Additionally, we make use of residual connections between corresponding layers, similar to that of U-Net\cite{unet}.

\begin{figure}[H]
\centering
\begin{minipage}{4in}
    \includegraphics[width=4in]{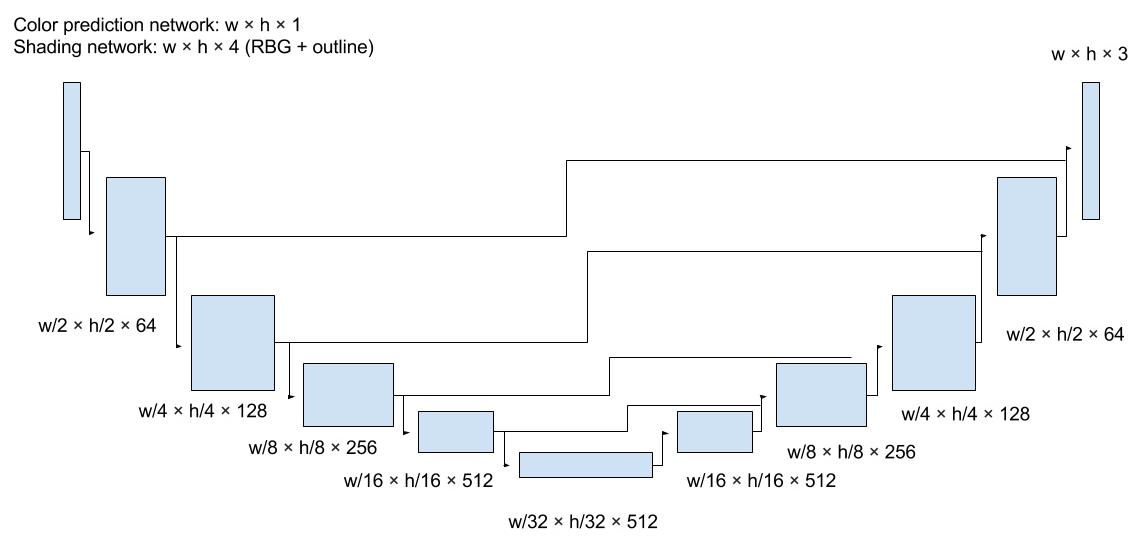}
\end{minipage}
\caption{Convolutional network structure. Arrows represent 5x5 convolutional filters, with a stride of 2.}
\end{figure}

\subsection{Tandem Structure}

We propose that a finalized piece of natural-looking artwork can be generated, given an image outline and a color scheme containing info on the general palette of an image. We also propose that if the color scheme is simple enough, it can be inferred to a reasonable degree from the image outline. The details on the color scheme will be discussed in the next section.

Accordingly, our setup contains two convolutional networks, as detailed in the previous section. The shading network maps from an outline ($W \times H \times 1$) and a color scheme ($W \times H \times 3$) to a final image ($W \times H \times 3$), while the color prediction network maps from the outline ($W \times H \times 1$) to a color scheme ($W \times H \times 3$). The shading network makes use of adversarial loss, while we found the color prediction network to perform better with a pure L2 loss.

These two networks serve a twofold purpose. Using only the shading network, a line image accompanied by a simple color scheme (such as messy scribbles) can be converted to a final piece. By introducing the color prediction network, the color scheme itself can be inferred, and converted into a final artwork through the shading network.

\begin{figure}[H]
\centering
\begin{minipage}{5in}
    \includegraphics[width=5in]{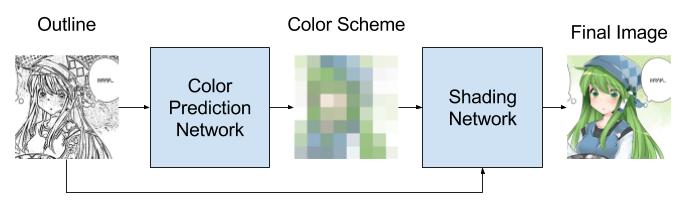}
\end{minipage}
\caption{Tandem structure for outline colorization.}
\end{figure}

\subsection{Adversarial Training}

A common framework in recent image generation advances is the generative adversarial network. It defines a custom loss function in the form of a discriminator network, that is learned in tandem with the generator. This discriminator maps from images to a scalar, returning a number close to 1 if it believes the given image is real, and a number close to 0 if it believes the image to be fake. In our network, we represent the discriminator as a stack of convolutional layers, along with one fully-connected layer at the end to map a feature matrix to the final probability.

\begin{figure}[H]
\centering
\begin{minipage}{5in}
    \includegraphics[width=5in]{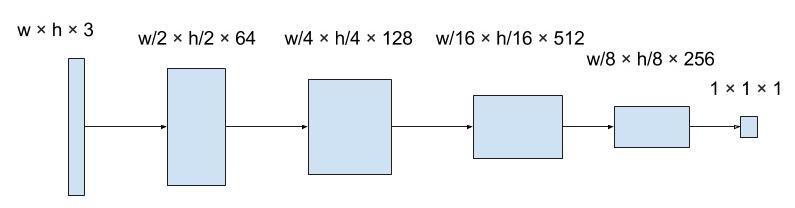}
\end{minipage}
\caption{Discriminator structure. Black arrows represent 5x5 convolutional filters, while the last white arrow is a fully-connected layer.}
\end{figure}

During each training iteration, we train the discriminator by feeding it both real and generated images along with their corresponding binary labels. We then train our generator by backpropagating through the discriminator, and attempting to convince it that our generated images are real.

To improve stability, we add a more traditional similarity metric of L2 distance to our loss function. This forces the generator to produce images that not only look real in the eyes of the discriminator, but also be close to the original in terms of pixelwise distance.

An important thing to note is the notion of a fully-convolutional network. Since convolutional layers are filters that are slid across the dimensions of an image, they can be reused regardless of an image's dimensions. This allows us to train entirely on 256x256 examples, but still colorize bigger images. The only non-convolutional aspect of our network is the final layer of the discriminator, and after training we simply discard the discriminator entirely.

\newpage
\section{Processing}

To our knowledge, there is no readily available dataset of line art and its corresponding final images. Creating such a dataset would require large amounts of manual work. Instead, we present a method of processing final artwork into outlines and color schemes, our network can learn from.

Producing an outline is the simpler of the two tasks. We run OpenCV's\cite{opencv} edge detection algorithm on the final images, with parameters adjusted in the style of real line art. Outlines generated this way perform well in terms of extracting shapes and borders. In heavily textured images, however, unwanted marks can appear.

The color scheme is more complicated. While some colors can be predicted purely from line art, such as skin, not enough context is present for others, such as clothes. To account for this, we present a method of representing general color locations (ex: bottom left is red, top right is blue) to account for this uncertainty.

It's important that the color scheme does not contain too much information, or the network would become too dependent on it. During training, the color scheme perfectly matches the desired final image, In practice, however, the color scheme will be provided by either a human or the color-choosing network, both of which are prone to error. Therefore, the network must be able to utilize messy and inaccurate color schemes, and still produce natural looking artwork.

To account for this need, we process the original artwork in a way that removes detail and adds noise. We repeatedly select 10x10 patches of the original image at random, and replace the contents with white. Patches are used instead of pixels, since individual pixels are likely to be correlated with those nearby. Removing entire regions of the image will likely remove information altogether. Finally, a large blur is applied to the whole image.

Once training is done, and we are putting it into practice, it is beneficial to make use of all information given. Therefore, we do not remove the patches, and instead scale down every pixel by its expected value. This patch removal method can be compared to applying dropout\cite{dropout} on the input image, treating patches as pseudo-independent variables.

\begin{figure}[H]
\centering
\begin{minipage}{1.1in}
    \includegraphics[width=1.1in]{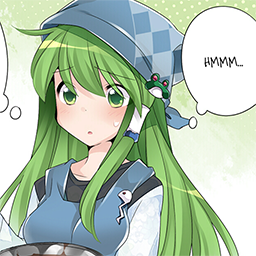}
\end{minipage}
\begin{minipage}{1.1in}
    \includegraphics[width=1.1in]{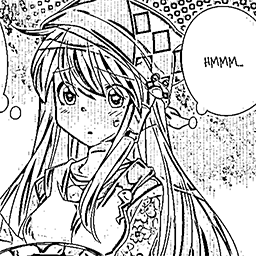}
\end{minipage}
\begin{minipage}{1.1in}
    \includegraphics[width=1.1in]{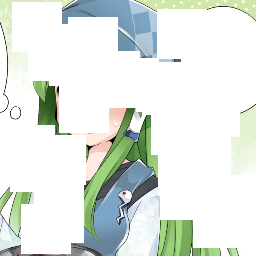}
\end{minipage}
\begin{minipage}{1.1in}
    \includegraphics[width=1.1in]{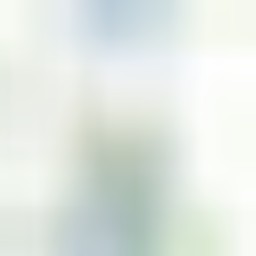}
\end{minipage}
\caption{Left to right: original, outline, color scheme with patches removed, color scheme after blur.}
\end{figure}

\newpage
The second network, which predicts colors from outlines, has a harder job. Directly mapping from outlines to their processed color schemes leads to poor results. To combat this, we adjust the second network's output. Rather than directly predicting every pixel's color, we split the image into a grid of 16px by 16px blocks. The second network is then tasked with assigning colors to these blocks, rather than the individual pixels within them.
By simplifying the second network's generation job, we allow it to focus on recognizing features, rather than producing intricate details. This works well as our first network has learned to expect a color scheme containing messy information.

\begin{figure}[H]
\centering
\begin{minipage}{1.5in}
    \includegraphics[width=1.5in]{original.png}
\end{minipage}
\begin{minipage}{1.5in}
    \includegraphics[width=1.5in]{outline.png}
\end{minipage}
\begin{minipage}{1.5in}
    \includegraphics[width=1.5in]{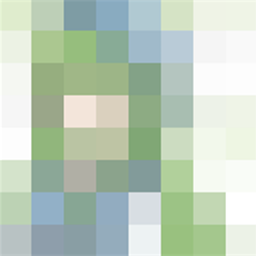}
\end{minipage}
\caption{Left to right: original, outline, desired output of color-prediction (second) network.}
\end{figure}

\newpage
\section{Experiments}

\footnotetext{Online demo available at http://color.kvfrans.com}

To test our setup, we build a set of ~10,000 manga/anime style illustrations from the online image database Safebooru. We select images depicting a single prominent human, according to user-assigned tags. Due to memory constraints, we train on 256x256 images with a batchsize of 4.

First, we test our shading network alone by generating outlines and color schemes from a separate validation set, and qualitatively inspecting the results. Images generated in this way, with perfect outlines and color schemes, show close resemblance to the originals. Interestingly, the generated images show a watercolor-like artistic style, rather than the usual hard colors.

\begin{figure}[H]
\centering
\begin{minipage}{5in}
    \includegraphics[width=5in]{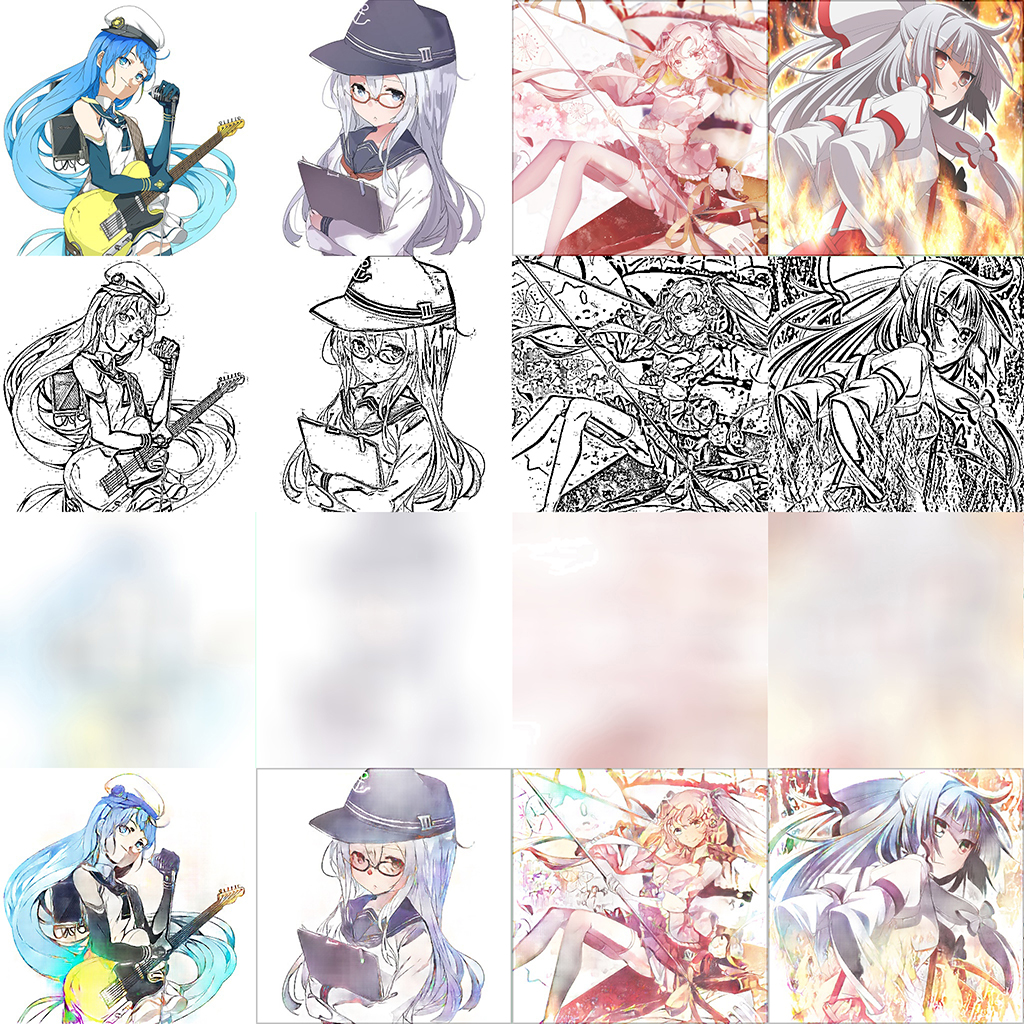}
\end{minipage}
\caption{Top to bottom: original images, outlines, color schemes, generated images}
\end{figure}

Secondly, we test our shading network's generalization ability by passing in hand-drawn color schemes, intentionally created to be messy. We frequently use only a few colors, go out of bounds, and miss areas completely. The network shows flood-fill like properties, preferring to use outlines as color borders. While hues are based mainly on the provided color scheme, color intensities appear to be mainly inferred from the outlines. We can also see accurate placement of shadows, even though the color black was never provided in the color schemes.

\begin{figure}[H]
\centering
\begin{minipage}{4in}
    \includegraphics[width=4in]{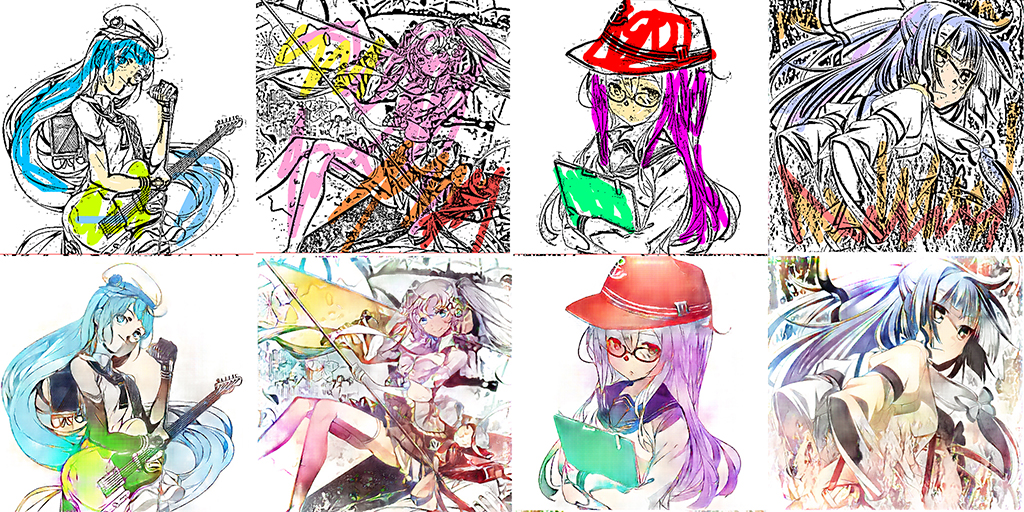}
\end{minipage}
\caption{Top: outlines + messy color schemes. Bottom: generated images}
\end{figure}

We also take a closer look at the shading network by using a completely white color scheme. With no hint of colors at all, the network is still able to create features such as shadows and highlights. In addition, common colors such as skin still appear.

\begin{figure}[H]
\centering
\begin{minipage}{4in}
    \includegraphics[width=4in]{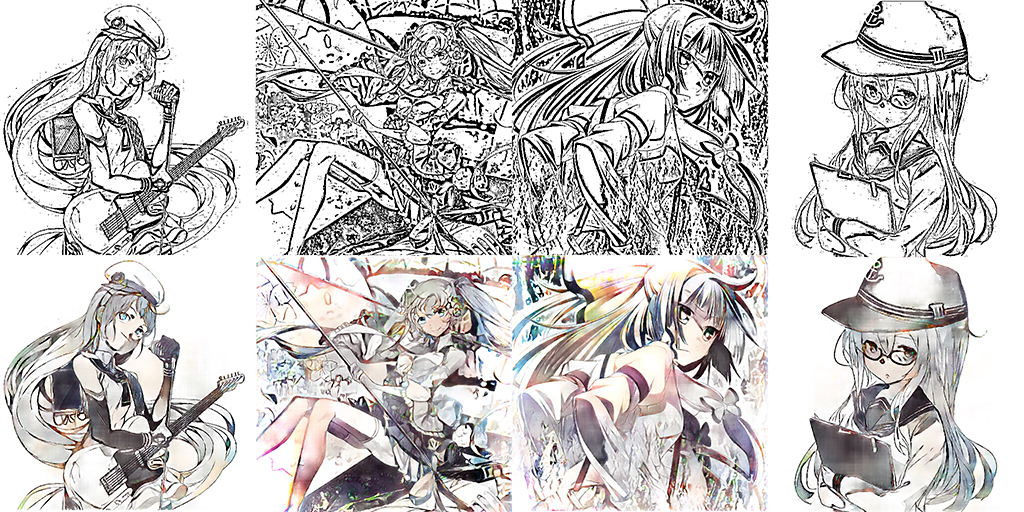}
\end{minipage}
\caption{Top: outlines + pure white color scheme. Bottom: generated images}
\end{figure}

Finally, we test the color prediction network by predicting color schemes from validation set outlines, and passing both into the shading network. The images produced are quite sharp: when examined individually, many details have been added, such as more distinct highlights and shadows. However, the network prefers to output similar-styled color predictions, leading to results lacking in diversity.

We compare this tandem method to a network that attempts to map directly from outlines to fully-colored images. The direct network creates images that lack the clear lines and shadows. Additionally, colors are placed somewhat randomly, often bleeding into other areas of the image. The tandem network provides sharper lines and constrains colors into stricter boundaries.

\begin{figure}[H]
\centering
\begin{minipage}{4in}
    \includegraphics[width=4in]{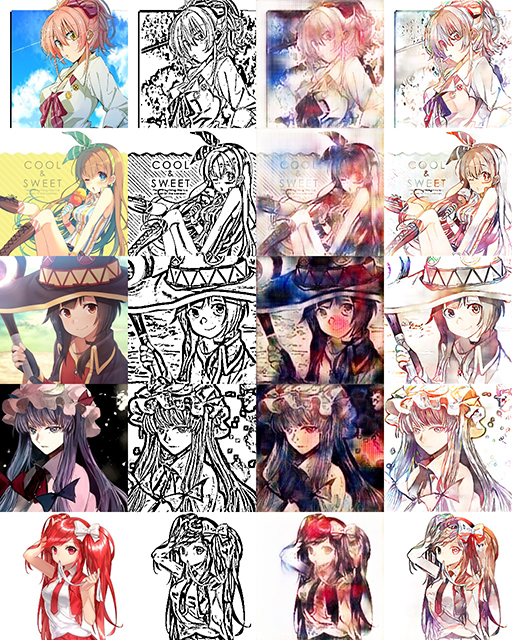}
\end{minipage}
\caption{Left to right: original images, outlines, images from direct method, images from tandem method}
\end{figure}

\section{Conclusion}

We present a tandem setup of two conditional adversarial networks to colorize and shade in line images. We introduce processing techniques to train the networks effectively by limiting information passed between the two. Lastly, we show natural-looking artwork can be generated from both messy, user-given color schemes and from color schemes predicted from the outlines. Future work could involve a color prediction network structured to encourage output diversity.

\end{document}